\pdfoutput=1

\documentclass[11pt]{article}

\usepackage{ACL2023}
\usepackage{listings}
\usepackage{xcolor}
\usepackage{amsmath}

\lstdefinelanguage{json}{
    basicstyle=\ttfamily\small,
    showstringspaces=false,
    breaklines=true,
    frame=single,
    stringstyle=\color{purple},
    keywordstyle=\color{blue}\bfseries,
    morestring=[b]",
    morekeywords={:,}, 
}

\usepackage{times}
\usepackage{latexsym}
\usepackage{graphicx}
\graphicspath{{./images/}}

\usepackage[T1]{fontenc}

\usepackage[utf8]{inputenc}

\usepackage{microtype}

\usepackage{inconsolata}

\usepackage[utf8]{inputenc} 
\usepackage[T1]{fontenc}    
\usepackage{hyperref}       
\usepackage{url}            
\usepackage{booktabs}       
\usepackage{amsfonts}       
\usepackage{nicefrac}       
\usepackage{microtype}      
\usepackage{xcolor}         
\usepackage{graphicx}
\usepackage{float}

 \usepackage{tabularray}

\title{Transfer of Structural Knowledge from Synthetic Languages}

\author{%
  Mikhail Budnikov\\
  School of Computer Science \& Engineering\\
  Constructor University\\
  Bremen, Germany 28359 \\
  \texttt{mbudnikov@constructor.university} \\
  \And
  Ivan Yamshchikov\\
  CAIRO\\
  THWS\\
  Würzburg, Germany 97082\\
  \texttt{ivan.yamshchikov@thws.de}\\
}

\begin{document}
\maketitle
\begin{abstract}
This work explores transfer learning from several synthetic languages to English. We investigate the structure of the embeddings in the fine-tuned models, the information they contain, and the capabilities of the fine-tuned models on simple linguistic tasks. We also introduce a new synthetic language that leads to better transfer to English than the languages used in previous research. Finally, we introduce Tiny-Cloze Benchmark — a new synthetic benchmark for natural language understanding that is more informative for less powerful models. We use Tiny-Cloze Benchmark to evaluate fine-tuned models in several domains demonstrating that fine-tuning on a new synthetic language allows for better performance on a variety of tasks.
\end{abstract}

\section{Introduction}

Large language models (LLMs) are becoming increasingly powerful and useful. However, the role of data properties in model training and what exactly models learn from the training data remains to a large extent out of the scope of most LLM papers. Yet surprisingly pre-training a model on a simple algorithmic task can lead to improvements in natural language modelling \citep{min2023recent}. Such insights can be used to improve the construction of data sets for language models. Therefore, exploring the mechanisms of knowledge transfer is an important open question. 

Scaling language models is a popular way to improve their performance\footnote{For a detailed review of the other ways to increase LLM generalization potential we address the reader to \citet{budnikov2024generalization}}. However, as the detailed analysis in \citet{villalobos2022will} shows, the amount of data, especially high-quality text data, is limited and will become the main bottleneck in the coming years.

Such circumstances motivate research into more data-efficient learning algorithms and a better understanding of the mechanisms of generalization and transfer learning \citep{surkov2024vygotsky}. 
After all, humans are exposed to orders of magnitude less data than modern frontier models, yet demonstrate strong performance across many domains and outperform machines in some areas, even considering recent algorithmic advances.

Inspired by this, \citet{huebner2021babyberta} demonstrate that training RoBERTa \citep{liu2019roberta} on language acquisition data, together with some tweaks to model architecture and training, leads to 6000$\times$ gains in data efficiency. Similarly, \citet{eldan2023tinystories} achieve significant model compression while retaining the ability to produce fluent and coherent English by using a generated dataset of stories for children, i.e. with small vocabulary and simple plots. And \citet{gunasekar2023textbooks} find that filtering for or generating data with higher educational value is also very helpful.

Thus, there is a growing body of evidence that the choice of data matters a lot and simply scraping the data from the web is suboptimal. However, there is a limited understanding of what properties of the data are important in different training stages.
\citet{papadimitriou2020learning} show that pre-training the LSTM \citep{hochreiter1997long} on structured but not linguistic data, such as MIDI music, Java code, or even nested parentheses, reduces its perplexity when testing on Spanish. \citet{sinha2021masked} find that removing all word order information from the pre-training phase does not significantly affect the final performance, given a fine-tuning phase with the correct word order. \citet{krishna2021does} sample the training data from a completely artificial language consisting of random n-grams and observe that pre-training objectives that require processing this information somehow, such as copying sentences in the right order, still improve the performance of the model on summarization tasks compared to a randomly initialized version. 

\begin{figure}[H]
\includegraphics[width=8cm]{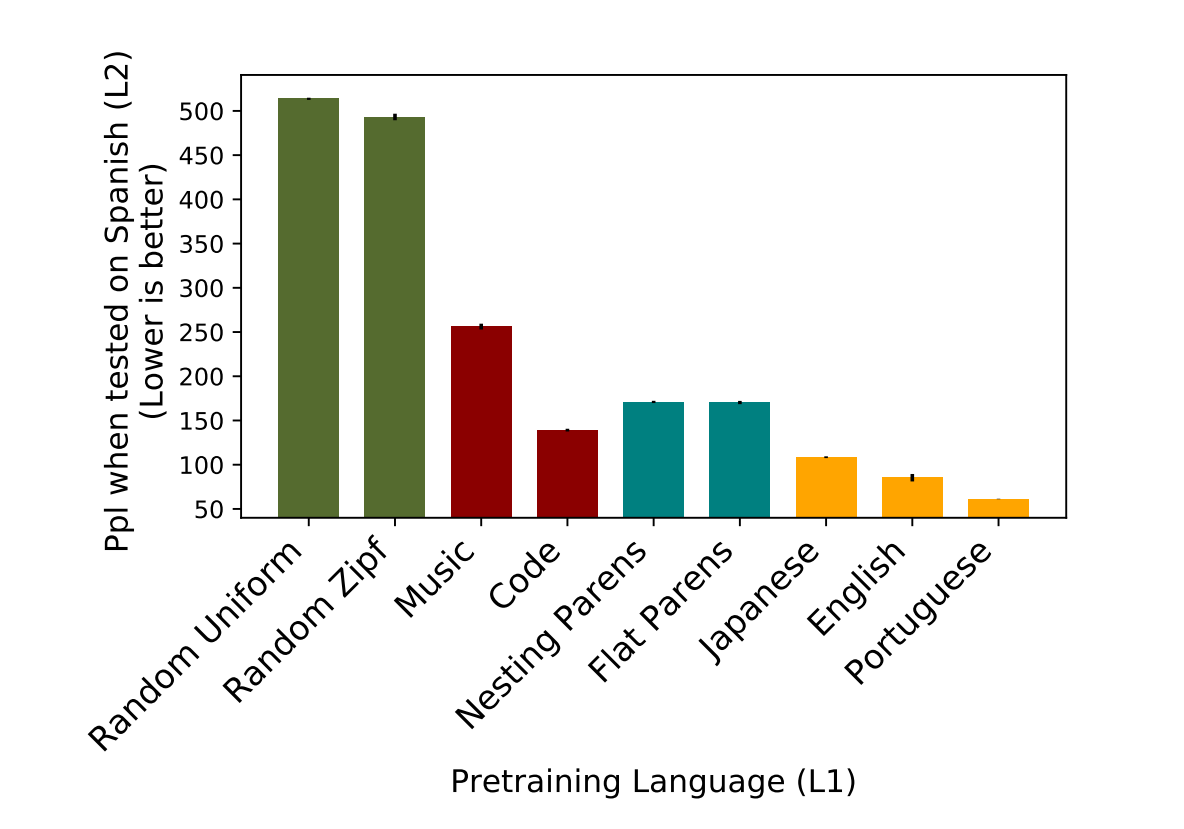}
\centering
\caption{Perplexity on various types of input \citep{papadimitriou2020learning}.}
\label{fig:obs}
\end{figure}

However, research in this area currently tends to focus on reporting surprising observations rather than explaining them. \citet{papadimitriou2020learning} illustrate such observations using Figure \ref{fig:obs}.

This work attempts to build up on those observations and make a small further step studying the mechanisms of transfer learning.

In this paper, we address the following research questions:
\begin{itemize}
  \item How do different synthetic pre-training datasets influence the complexity and transfer performance of language models on English tasks?
  \item What structural properties of the learned embeddings reflect the characteristics of the pre-training data?
  \item To what extent do synthetic pre-training languages affect the encoding of linguistic features in embeddings, as measured by linear probes?
\end{itemize}

Our contributions are summarized as follows:
\begin{itemize}
  \item We introduce a new synthetic language, \texttt{flat\_shuffle}, combining shuffle-based and bracket-based patterns, and compare it with existing synthetic datasets.
  \item We propose a transfer-learning-based measure to quantify language complexity and similarity via fine-tuning dynamics.
  \item We analyze the structure of embeddings through singular value spectra and clustering to assess their effective dimensionality.
  \item We employ linear probes to examine the linguistic information captured in embeddings fine-tuned on different synthetic languages.
  \item We present the Tiny-Cloze Benchmark, a synthetic NLU benchmark generated with GPT-4, demonstrating the benefits of synthetic pre-training.
\end{itemize}

First, as can be seen in the diagram above, different pre-training datasets, even if they all are unrelated to the target task, lead to different final performance. This suggests that some datasets are inherently more complex or more similar to the target language. We introduce a new synthetic "language" by combining ideas from the previous work and use it, as well as two existing synthetic datasets, to pre-train the models. We then fine-tune them on English using three different fine-tuning pipelines. We also provide an algorithm to assess the impact of the pretraining data on the resulting model parameters. 

Second, since one of the settings for transfer learning involves fine-tuning only the embeddings, they are the natural target for investigation. We investigate the structure of the learned embeddings, namely the spectrum of their singular values to understand the effective dimensionality of the data, and the KMeans clustering of to check how uniformly the embeddings are distributed. To check what information is contained in the embeddings, we train linear probes to predict certain features of the words given their embeddings. Linear probes are a popular interpretability technique, but to our knowledge they have not been used to study the embeddings of models pre-trained on different datasets and fine-tuned to the same task.  

Finally, we evaluate the performance of these models in natural language understanding. Since existing NLU datasets such as GLUE \citep{wang2018glue} and MMLU \citep{hendrycks2020measuring} are designed for more powerful models, we use GPT-4 \citep{openai2023gpt4} to generate a similar benchmark consisting of 12 different subtasks\footnote{To facilitate reproducibility and further research, we publish our code and data \url{https://github.com/msh2481/language_transfer}}. 

\section{Related work}
 
 One way of understanding the pre-training of language models is that we transfer some linguistic knowledge from a task with lots of available data to a downstream task \citep{han2021pre}. The recent findings suggest that this is not the only relevant effect, and sometimes not even the most important one. 

\citet{papadimitriou2020learning}, mentioned above, pre-trained an LSTM on structured but not linguistic data and found that adapting such a model to Spanish by fine-tuning only its input and output embeddings gave better perplexity than starting with a randomly initialised model. Their results and experimental setup established a framework that has been used in subsequent work, including this one. \citet{ri2022pretraining} improved these results replacing LSTM with a Transformer and changing the synthetic pre-training languages. \citet{papadimitriou2023injecting} used GPT-2 \citep{radford2019language}. \citet{chiang2022transferability} introduce a family of Shuffle languages. \citet{artetxe2019cross} used a similar technique to combine a task-specific corpus in English with a corpus in the target language unrelated to the task.

Such transfer also works in the opposite direction, from natural language to other domains. \citet{lu2021pretrained} get performance comparable to training from scratch on different modalities by fine-tuning only the input and output embeddings of the pre-trained GPT-2. They also try different fine-tuning approaches, tuning the layer norm parameters and the last Transformer block in addition to the input and output embeddings.

\citet{mehta2021empirical} show that pre-training moves the model parameters into a flat basin of the loss landscape and suggest it as a reason why pre-trained models are less prone to catastrophic forgetting during fine-tuning. \citet{neyshabur2020being} also observe this and also show that models fine-tuned from the same checkpoint stay in the same basin. However, past data alone is almost never enough to predict unseen data, unless one makes some assumptions, i.e. "inductive bias". A useful inductive bias can be injected into the model by pre-training on data that has it. \citet{mccoy2020universal} use pre-training on natural languages with certain properties by model-agnostic meta-learning \citep{finn2017model} to find which biases are needed to quickly acquire these languages. \citet{wu2021lime} design synthetic datasets requiring deduction, induction, and abduction and pre-train on them to extract inductive bias for general mathematical reasoning. \citet{lindemann2023injecting} pre-train models to simulate finite state transducers given their description and achieve better generalization in NLP tasks with similar structure.

\section{Synthetic Languages}

In this paper we report a series of experiments with several synthetic languages. Following hyperparameter choices from \citet{papadimitriou2023injecting}, for each of the languages described below, we use a sequence length of $512$, a vocabulary
size of $500$, and generate $2 \cdot 10^{6}$ sequences so the total size of the dataset is approximately $10^{9}$ tokens in each case.
  
We focus on three synthetic languages: \texttt{nested}, the k-Dyck nested bracket language; \texttt{flat}, the shuffle Dyck language with no nesting; and \texttt{flat\_shuffle}, a block-wise shuffled variant of the flat language.

\subsection{\texttt{nested}}

\citet{papadimitriou2020learning} used a stack-based grammar to generate sequences, where each token occurs twice and two pairs of tokens either do not intersect or one is nested in another. In other words, a balanced bracket sequence with multiple types of brackets.

\citet{ri2022pretraining} suggested using different tokens for opening and closing brackets, and found improved performance. We chose to implement this version, and save a synthetic language with $250$ tokens for open brackets and $250$ tokens for closing ones.

Tokens are generated sequentially, and on each step, a random decision is made whether to open a new bracket or to close an existing one. If the stack of open brackets is empty or there is not enough space before the end of the sequence, there is only one
option. In other cases, an opening bracket is chosen with a probability of $0.4$, and then the type of bracket is sampled uniformly.

\definecolor{green}{RGB}{50,200,50}
\definecolor{blue}{RGB}{50,50,200}
\definecolor{yellow}{RGB}{200,200,0}
\definecolor{purple}{RGB}{150,0,150}
\definecolor{p1}{RGB}{200,0,100}
\definecolor{p2}{RGB}{166,0,133}
\definecolor{p3}{RGB}{133,0,166}
\definecolor{p4}{RGB}{100,0,200}
\definecolor{g1}{RGB}{150,200,150}
\definecolor{g2}{RGB}{183,200,116}
\definecolor{g3}{RGB}{216,200,83}
\definecolor{g4}{RGB}{250,200,50}

\newcommand{\bop}[2]{\textcolor{#1}{\texttt{<#2}}}
\newcommand{\bcl}[2]{\textcolor{#1}{\texttt{#2>}}}

Example word from \texttt{nested}:

\bop{green}{23} \bop{yellow}{42} \bop{blue}{15} \bcl{blue}{15} \bcl{yellow}{42} \bcl{green}{23} \bop{purple}{56} \bcl{purple}{56}

\subsection{\texttt{flat}}
This language is similar to the previous one. The only difference is that the nesting property can be violated.

In terms of sampling, it means that when a bracket should be closed, now there is more than one possibility. We select the bracket to close uniformly from all currently open ones.

Example word: 

\bop{green}{23} \bop{yellow}{42} \bop{blue}{15} \bcl{yellow}{42} \bcl{green}{23} \bop{purple}{56} \bcl{blue}{15} \bcl{purple}{56}

\subsection{\texttt{flat\_shuffle}}
The \texttt{flat\_shuffle} language extends \texttt{flat} by partitioning bracket type IDs into contiguous blocks of size eight, such that each segment of 16 tokens is sampled exclusively from one block, yielding a permutation of those brackets within the segment. While the languages described above are each based on a single rule, this extension introduces additional structure to the data, which we hypothesize can improve model performance.

We suggest to use an idea of \texttt{shuffle} languages from \citet{chiang2022transferability} as an extra pattern because it was orthogonal to the bracket balancing essence of the previous datasets. The combined dataset is based on \texttt{flat}, but
each consecutive group of $16$ tokens has a range of $8$ bracket types assigned to it, and all brackets on this segment are sampled only from these types. That is, each such group is a permutation of the corresponding brackets.

It adds two interesting properties to the task of next token prediction. First, in the middle of the line the model has to look at previous tokens to guess the range of bracket types to predict the next token. Second, the model can guess increasingly more accurately by remembering which tokens were already used if we are close to the end of the permutation. In particular, the last token in each permutation can be predicted with certainty. Surprisingly, even small Transformer models were able to capture this pattern and
indeed predicted the last token with close to zero loss.

Example word (purple and green parts represent two blocks, $[16, 20)$ and $[36, 40)$): 

\bop{p1}{16} \bop{p3}{18} \bcl{p1}{16} \bop{p2}{17} \bop{p4}{19} \bcl{p3}{18} \bcl{p2}{17} \bop{g3}{38} \bcl{g3}{38} \bop{g1}{36} \bcl{p4}{19} \bop{g4}{39} \bcl{g4}{39} \bop{g2}{37} \bcl{g1}{36} \bcl{g2}{37}

\section{Methodology}

Some languages, both synthetic and natural, are more complex than others. For example, it is much easier to understand the concept of balanced bracket sequences than to learn Assyrian language. Moreover, some languages can be understood more easily if the learner already knows another language. For instance, humans need less effort to learn a language from the same language family, and large language models can be fine-tuned for a similar downstream task using much less data than was used for their pre-training.

One approach to formalize this intuition of
complexity and similarity is the Chomsky hierarchy of languages \citep{chomsky1956three}. It formally defines several classes of grammars, each one strictly more general than the previous one, and the properties of these classes are very well understood.
For example, \texttt{nested} is a context-free language, while \texttt{flat} is context-dependent. However, for languages from the same class, we need some other tool to find more fine-grained
differences. We propose a transfer-learning based approach to quantify those.

An important observation is that transfer learning between languages is not symmetric, and it allows us to estimate both (relative) complexity and similarity of two languages. If languages are similar, transfer learning should go well in both
directions. However, if one language is more complex than another, at least in the sense of having strictly more patterns, one would expect transfer learning to be much easier from the hard language to the easy one. So, assuming that we have some operationalization $f(A, B)$ of "difficulty of transfer learning from language $A$ to language $B$", we can take $\frac{1}{2}(f(A, B) + f(B, A))$ to mean dis-similarity of $A$ and $B$ and $\frac{1}{2}(f(B, A) - f(A, B))$ to mean complexity of $A$ relative to $B$.

Our proposed way to operationalize this notion of "difficulty" is to just use perplexity of the model pre-trained on language $A$ and then fine-tuned to language $B$, with some of the weights frozen. By varying the subset of the weights allowed to be fine-tuned we can get a more complete picture, i.e. for some pair of languages just tuning the embeddings might already be enough, which would mean that they share most of the structure.

A more practically-oriented way to compare synthetic languages is to see which of them better prepare models to learning English. To test this we take models pre-trained on each of the synthetic languages, fine-tune them to English, and check their language understanding capabilities with cloze questions.

Finally, we study the structure of the embeddings in terms of effective dimensionality and number of clusters, and then explore what English word features are learned by models fine-tuned from each of the synthetic languages.

\section{Experiments}

For all experiments, we used the TinyStories-8M model
\citep{eldan2023tinystories}. 

\subsection{Transfer Learning}

We used three levels of trainable weight subsets:

\textbf{E}: Only input and output embeddings are tuned;

\textbf{EL}: \textbf{E} plus the affine parameters of LayerNorms;

\textbf{ELT}: \textbf{EL} plus the entire last Transformer layer;

For pre-training, we waited until convergence that was close to the theoretical lower bounds of loss or just long stagnation, which took $40$K to $100$K steps. The batch size was $8$, and the sequence length was $512$ tokens, so we used $160$M to $400$M tokens for pre-training. For fine-tuning,
at each stage, we used a fixed amount of $12500$ steps, the batch size was again $8$, and the sequence length was $512$ for bracket datasets and $128$ for English (TinyStories), which means $51$M and $13$M tokens correspondingly. The learning rate was $10^{-3}$ for pre-training and $[10^{-2}, 2 \cdot 10^{-2}, 10^{-3}]$ for the three stages of fine-tuning.

Table \ref{tab:exp} below presents the results of fine-tuning in both directions on certain pairs of languages.

\begin{table*}[h]
\centering
\begin{tblr}{
  cell{1}{1} = {c=2}{},
  cell{1}{3} = {c=4}{},
  cell{1}{7} = {c=4}{},
}
Language pair &    & L1 $\rightarrow$ &       &        &         & L2 $\rightarrow$ &       &        &         \\
L1          & L2 & L2E                             & L2EL & L2ELT & L2Full & L1E & L1EL & L1ELT & L1Full \\
\hline
\texttt{nested}        & \texttt{flat}          & 4.4          & 4.1           & 4.1            & 3.8             & 3.5          & 3.3           & 3.3            & 3.3             \\
\hline
\texttt{flat}          & \texttt{flat\_shuffle} & 2.5          & 2.4           & 2.2            & 2.0             & 3.8          & 3.8           & 3.7            & 3.8             \\
\hline
\texttt{flat\_shuffle} & English                & 2.4          & 2.3           & 2.0            & 1.2            & 2.8          & 2.6           & 2.1            & 2.0             \\
\hline
\texttt{nested}        & English                & 2.8          & 2.7           & 2.4            & 1.2             & 3.8          & 3.5           & 3.3            & 3.3             \\
\hline
\texttt{flat}          & English                & 2.7          & 2.6           & 2.4            & 1.2             & 4.3          & 4.2          & 3.8            & 3.8             \\
\hline
\end{tblr}
\caption{Pretraining on L1 and transferring to L2 and vice versa. Values are negative log-likelihoods in nats. "E", "L", and "T" indicate which layers were fine-tuned and stand for embeddings, LayerNorms, and (the last) Transformer block respectively. Columns ending in "Full" correspond to models trained from scratch on the respective language (i.e., no synthetic pretraining). We use the absolute difference of $0.2$ nats per token as a threshold for "close performance".}
\label{tab:exp}
\end{table*}

The first row shows that \texttt{flat} is more complex than "nested". The second row demonstrates that \texttt{flat\_shuffle}
is more complex than \texttt{flat}. Indeed, fine-tuning in the direction \texttt{flat\_shuffle} $\to$ \texttt{flat} $\to$ \texttt{nested} achieves relatively good performance already with the first stage of fine-tuning. The other experiments
show that English is more complex than all synthetic languages used here, but it is also quite different, as the model needs more flexibility to adapt from English to \texttt{flat} or \texttt{flat\_shuffle}.

\subsection{Cloze Tests}

To assess how well the models understand language in general, a different benchmark is needed. Since the models studied are too small for reliable question answering, reasoning, and other high-level cognitive skills, the test should be as simple as possible, ideally just measuring perplexity on some texts. There are existing datasets for natural language understanding, such as GLUE
\citep{wang2018glue} and MMLU \citep{hendrycks2020measuring}, but they focus on more complex tasks.

Instead, we used GPT-4 \citep{openai2023gpt4} to generate Tiny Cloze Benchmark — a set of cloze\footnote{\url{https://en.wikipedia.org/wiki/Cloze_test}} infilling questions in simple English. There are the following $12$ subtasks, each with $10$ cloze questions: 'synonyms and antonyms' — the model chooses one of two antonyms to correctly fill the gap in the sentence; 'Logical relations' —  the model chooses a correct conjunction between two parts of the sentence; 'Subject-verb agreement' — the model chooses one of two verbs that corresponds to the given subject in the sentence; 'Prepositions' — the model chooses a correct preposition in the sentence;  'Conjunctions' — a task similar to 'Logical relations' but with different conjunctions; 'Temporal understanding' — filling in a correct temporal conjunction; 'Spatial understanding' — filling in a word based on spatial understanding of the sentence; 'Quantitative reasoning' — filling in the number into the sentence; 'Emotions' — filling the correct emotional adjective into the sentence; 'Narrative understanding' — filling one noun relevant for the narrative sentence; 'Ethics' — filling a noun for an ethical statement. You can find detailed examples of the tasks in Appendix.

Each cloze question consists of a prompt with a cloze marker, a correct answer, and an incorrect answer. For each question, the difference between log-probabilities of the correct and incorrect answers is measured and then averaged across each subtask. We measure the difference in log-likelihoods rather than accuracy, because it provides more information per sample, which is important given the relatively small size of our benchmark.

Here is an example question from the temporal understanding subtask:

\texttt{[ "She ate breakfast \# she went to school", "before", "after", ] }

For each of the synthetic languages, we used two models, one in which only the embeddings were fine-tuned on English (E), and another with all three stages of fine-tuning applied (ELT). We compared them with the model of the same architecture (8M parameters) trained on English from scratch and also to a four times larger model with 33M parameters trained on English from scratch to see which metrics can be improved.

\begin{table*}[ht]
\centering
\begin{tabular}{|p{70pt}|*{8}{r|}}
\hline & \texttt{\scriptsize{nested E}} & \texttt{\scriptsize{nested ELT}} & \texttt{\scriptsize{flat E}} & \texttt{\scriptsize{flat ELT}} & \texttt{\scriptsize{flat shuffle E}} & \texttt{\scriptsize{flat shuffle ELT}} & \texttt{\scriptsize{scratch 8M}} & \texttt{\scriptsize{scratch 33M}}\\
\hline
    \scriptsize{synonyms and antonyms} & 0.24 & 0.18 & 0.22 & 0.13 & 0.31 & 0.25 & 0.25 & 0.28 \\ \hline
    \scriptsize{single - plural} & 0.08 & 0.15 & 0.19 & 0.50 & 0.03 & 0.33 & 0.58 & 0.71 \\ \hline
    \scriptsize{logical relations} & -0.08 & -0.30 & -0.44 & -0.18 & -0.13 & -0.08 & -0.04 & 0.09 \\ \hline
    \scriptsize{subject-verb agreement} & 0.54 & 0.45 & 0.46 & 0.36 & 0.14 & 0.26 & 0.83 & 0.98 \\ \hline
    \scriptsize{prepositions} & 0.43 & 0.52 & 0.51 & 0.53 & 0.40 & 0.48 & 0.94 & 1.12 \\ \hline
    \scriptsize{conjunctions} & 0.46 & 0.43 & 0.45 & 0.38 & 0.36 & 0.49 & 0.63 & 0.82 \\ \hline
    \scriptsize{temporal understanding} & -0.13 & -0.02 & -0.14 & 0.04 & 0.09 & 0.36 & 0.44 & 0.73 \\ \hline
    \scriptsize{spatial understanding} & 0.13 & 0.30 & 0.40 & 0.48 & 0.06 & 0.37 & 0.64 & 0.71 \\ \hline
    \scriptsize{quantitative reasoning} & -0.06 & 0.00 & -0.14 & -0.01 & -0.14 & -0.04 & -0.04 & -0.06 \\ \hline
    \scriptsize{emotions} & -0.08 & 0.03 & 0.05 & 0.07 & 0.20 & -0.01 & 0.61 & 0.77 \\ \hline
    \scriptsize{narrative understanding} & -0.07 & -0.04 & 0.03 & 0.07 & 0.04 & 0.04 & 0.17 & 0.27 \\ \hline
    \scriptsize{ethics} & 0.32 & 0.17 & 0.34 & 0.22 & 0.27 & 0.30 & 0.25 & 0.51 \\ \hline
\hline\scriptsize{{\textbf{{Average}}}} & 0.15 & 0.16 & 0.16 & 0.22 & 0.14 & 0.23 & 0.44 & 0.58 \\
\hline
\end{tabular}
\caption{Results on the Tiny-Cloze benchmark. Values show differences in log-likelihoods (in nats) between correct and incorrect answers. Fine-tuning on \texttt{flat\_shuffle} gives the highest average score across three synthetic languages.}
\label{tab:cloze}
\end{table*}

As shown in Table \ref{tab:cloze}, there are two interesting observations. First, models with all three stages of fine-tuning are better, predictably, than their counterparts having only the embeddings tuned, but this difference is more pronounced in
\texttt{flat} and \texttt{flat\_shuffle}. Second, Table \ref{tab:cloze} again shows the familiar pattern \texttt{nested} \textless \texttt{flat} \textless \texttt{flat\_shuffle} \textless \texttt{scratch},
which proves the superiority of the introduced \texttt{flat\_shuffle} dataset.

\subsection{Dimensionality and Clusters}
We hypothesize that models pretrained on more complex languages will exhibit a slower decay in the singular value spectrum, reflecting higher effective dimensionality required to encode richer structure.

The embedding dimension of the model used is $d = 256$, and human intuition, as well as many visualization techniques, work poorly for $256$-dimensional vectors. We hypothesize that the singular value spectrum reflects the encoding complexity of the pre-training language: languages requiring richer structure will exhibit a slower decay in the spectrum tail, indicating higher effective dimensionality in the embeddings. Therefore, we employ two quantitative approaches.

First, for an $n \times d$ matrix of embeddings $E$, we consider its singular values (after zeroing out the mean of each column), or equivalently, the spectrum of the covariation matrix $A = E^{T}E$. The motivation behind this is that if all
embeddings were contained in a $k$-dimensional subspace, and $E$ had a rank $k$, then only $k$ of the singular values would be nonzero. For real data, it is not the case, as all singular values are nonzero, but still, some directions have much larger variance than others, and the model is more likely to use features corresponding to those dimensions.

\begin{figure}[H]
\centering
\includegraphics[width=0.48\textwidth]{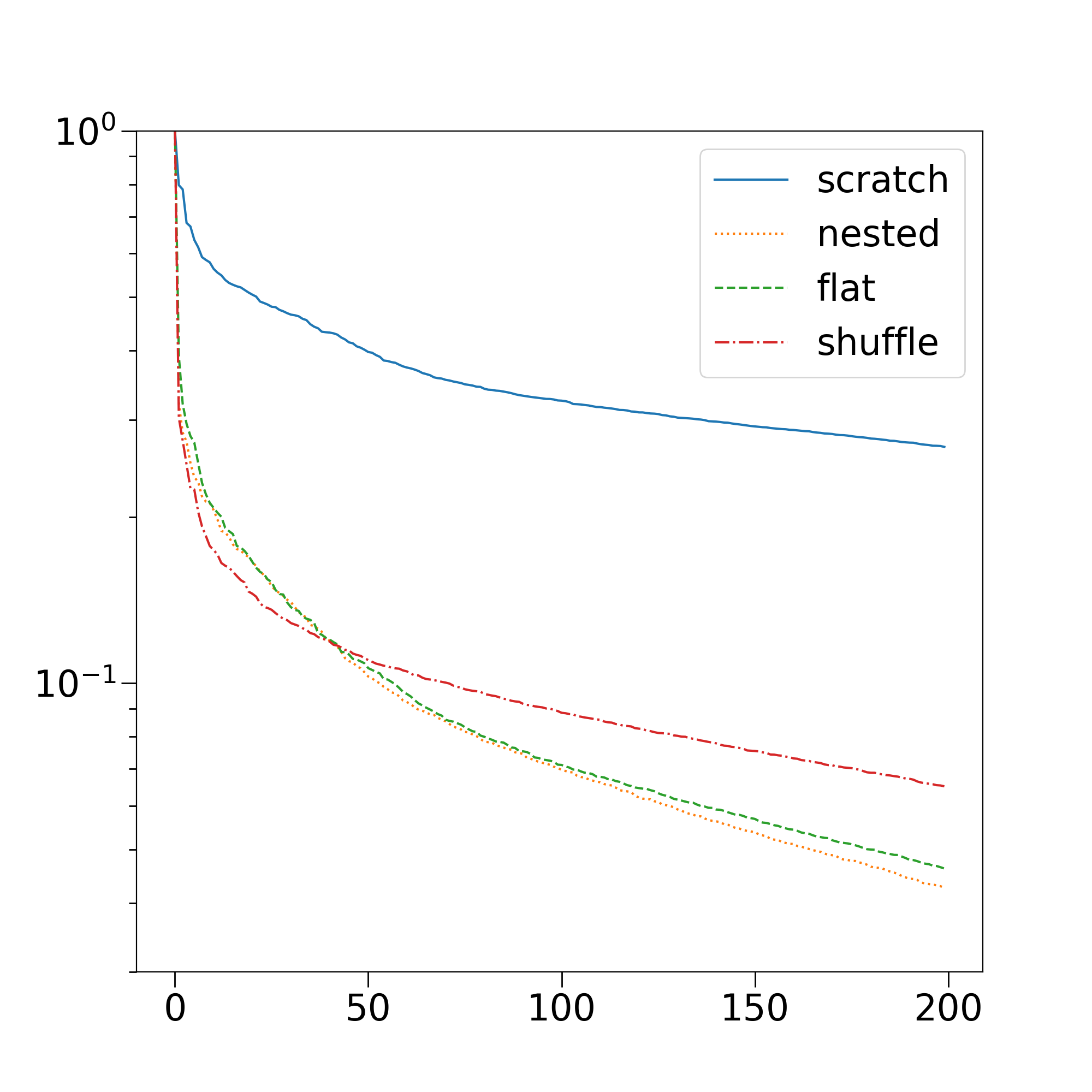}
\caption{Spectrum of bracket embeddings}
\label{fig:spectrum}
\end{figure}

As we see in Figure \ref{fig:spectrum}, in models pre-trained on synthetic datasets, the spectrum is dominated by the first few dimensions. In particular, before fine-tuning, most of the interesting information about brackets is described by
two axes: open-close and low-to-high bracket type id. While they learn more diverse features during fine-tuning on English, as described in the next sections, they still don't use the embedding space very efficiently. An interesting observation is how the tail of the spectrum behaves for models trained on different datasets: the spectrum of \texttt{flat} decays to zero slower than the one of \texttt{nested}, but the shape is similar, while the spectrum of \texttt{flat\_shuffle} crosses \texttt{flat} at some point and behaves more similarly to the spectrum of the model trained on English from scratch.

\begin{figure}
\centering
\includegraphics[width=0.48\textwidth]{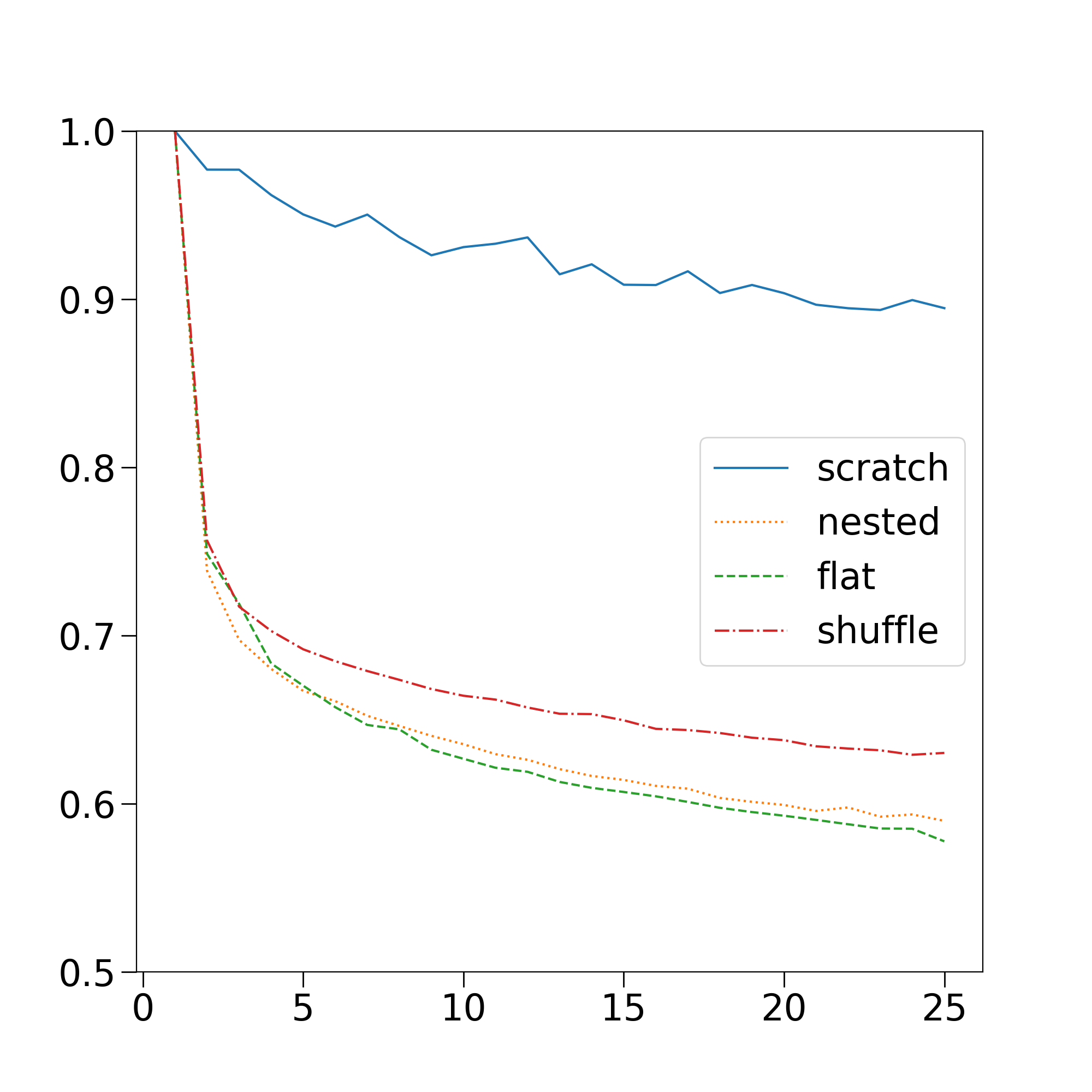}
\caption{Clustering of bracket embeddings}
\label{fig:clusters}
\end{figure}

Another interesting property is how the embeddings are clustered. To quantify it, we run k-means clustering for the embeddings varying the number of clusters and compare the plots of unexplained variance (Figure \ref{fig:clusters}). Again, after pre-training on a synthetic language, the models have only two clusters: open and close brackets, and even after fine-tuning, the first few splits explain the majority of variance. Looking at the tail behavior, we observe a similar pattern: English is followed by \texttt{flat\_shuffle}, then by \texttt{flat} and \texttt{nested}.

The \texttt{scratch} provides a reference for the embeddings on English. We can clearly see that \texttt{flat\_shuffle} embeddings are characteristically different from \texttt{flat} and \texttt{nested} embeddings both in terms of the spectrum and in terms of the clusters they form.

\subsection{Linear Probes for Word Features}

We train all probes on embeddings obtained from embedding-only (E) fine-tuned models to isolate the impact of embedding space structure.

Now that we know something about the structure of the embedding space, a natural question to ask is how this structure is used. In other words, what information about a word can one extract from the embedding of the corresponding token?

Preliminary experiments showed that clusters of features correspond to properties like "noun", "3rd person verb", "adjective or adverb", etc. We hypothesize that embeddings from more complex synthetic pretraining (e.g., \texttt{flat\_shuffle}) better capture such linguistic features. 
Consequently, we extract part of speech tags using NLTK. Given the limited vocabulary of TinyStories, capabilities of NLTK POS tagger should be good enough for our purposes.
Initially, there were more than $30$ unique tags in the dataset, but many of them were very rare. After filtering out all tags with less than $200$ occurrences, the following tags remained: CD — cardinal digit; IN — preposition or subordinating conjunction; JJ — adjective; NN - singular noun; NNP – proper noun; NNS – plural noun; RB – adverb;  VB – base form verb; VBD - past tense verb; VBG - gerund; VBN - past participle. We use this notation in  Table \ref{tab:prob}.

We added a feature indicating the frequency of the token in the training corpus because typically the direction with the most variance in the embedding space roughly corresponded to frequency. We also added another boolean feature that is one if the token starts with a whitespace and zero otherwise.

For each of the models and each of the features, we trained a ridge regression (for frequency) or a logistic regression (for all other variables, as they are Boolean) on $80\%$ of the embeddings and then evaluated their $R^{2}$ score or ROC-AUC on the remaining $20\%$. See Table \ref{tab:prob} for the results.

\begin{table*}[ht]
\centering
\begin{tabular}{ |p{70pt} |p{40pt} |p{40pt} |p{40pt} |p{40pt}| }
\hline
                 & \texttt{nested} & \texttt{flat} & \texttt{flat\_shuffle} & \texttt{scratch} \\
\hline
frequency        & 0.84            & 0.85          & 0.85             & 0.93             \\
\hline
start\_space     & 0.70            & 0.70          & 0.70             & 0.89             \\
\hline
pos\_tag\_CD     & 0.66            & 0.63          & 0.63             & 0.80             \\
\hline
pos\_tag\_IN     & 0.76            & 0.79          & 0.71             & 0.87             \\
\hline
pos\_tag\_JJ     & 0.60            & 0.58          & 0.60             & 0.73             \\
\hline
pos\_tag\_NN     & 0.63            & 0.62          & 0.63             & 0.76             \\
\hline
pos\_tag\_NNP    & 0.64            & 0.65          & 0.63             & 0.79             \\
\hline
pos\_tag\_NNS    & 0.67            & 0.67          & 0.68             & 0.84             \\
\hline
pos\_tag\_RB     & 0.69            & 0.63          & 0.64             & 0.84             \\
\hline
pos\_tag\_VB     & 0.71            & 0.69          & 0.68             & 0.79             \\
\hline
pos\_tag\_VBD    & 0.75            & 0.71          & 0.67             & 0.89             \\
\hline
pos\_tag\_VBG    & 0.71            & 0.70          & 0.73             & 0.89             \\
\hline
pos\_tag\_VBN    & 0.72            & 0.68          & 0.72             & 0.87             \\
\hline
\textbf{Average} & 0.70            & 0.68          & 0.68             & 0.84             \\
\hline
\end{tabular}
\caption{ROC-AUC for the linear probes on embeddings from models fine-tuned on English using embedding-only (E) mode.}
\label{tab:prob}
\end{table*}

All probes in all models perform better than random, so every model learns at least something related to these word features. The embeddings of the model trained on English from scratch predictably outperformed the others, but the quality of other
embeddings turned out to be on average the same. Perhaps the difference in effective dimension between the models is used not for these relatively simple single-word features, but for more complex ones.

\section{Conclusion}

We introduced a new synthetic language \texttt{flat\_shuffle}, and the model pre-trained on it was shown to outperform the models based on the languages from previous work.

Investigation of the structure of the embeddings leads to a hypothesis that the reason behind the superior performance of some synthetic languages is that they require more structured embeddings, which causes the intermediate layers to be adapted to work with such embeddings, and in turn allows effectively using a higher dimension subspace of the embedding space during fine-tuning, which gives more flexibility. 

Also, we haven't observed direct transfer of structure from synthetic languages to English, i.e. English tokens weren't splitted by the model into "opening" and "closing" ones. So it seems that models are working in a reservoir computing style where the computations for an unrelated task are adapted to the task at hand in arbitrary ways. At the same time, it means that the models are not strictly limited by the complexity or structure of the original task in transfer learning, and as long as they have enough complexity of computations, they can use it to adapt to the new task.

\section{Limitations}

The experiments reveal several interesting patterns about transfer learning between synthetic and natural languages. However, our approach has some important limitations.

First, we only used English as the target natural language. It would be interesting to see if the patterns we observed hold for other natural languages, especially those with different grammatical structures.

Second, even our most complex synthetic language, \texttt{flat\_shuffle}, was simple enough to be learned by a model with 8 million parameters. Perhaps with better synthetic data and correspondingly more capable models we would observe qualitatively new phenomena.

\section*{Ethics Statement}

This paper complies with the \href{https://www.aclweb.org/portal/content/acl-code-ethics}{ACL Ethics Policy}.



\newpage
\bibliographystyle{acl_natbib}
\bibliography{refs}


\appendix

\section{Appendix}

Here are the examples of the Tiny Cloze benchmark for particular tasks. One example for each task:
\begin{itemize}
    \item 'synonyms and antonyms': \\
    \texttt{"The box was incredibly light, almost as if it were $\#$.", "empty", "full"} 
    \item 'single plural': \\
    \texttt{They $\#$ to the store every Saturday.", "go", "goes"} 
    \item 'logical relations': \\
    \texttt{"The dog barked loudly, $\#$ everyone woke up", "and", "but"}
    \item 'subject-verb agreement': \\
    \texttt{"The dog in the yard $\#$ every morning", "barks", "bark"}
    \item 'prepositions': \\
    \texttt{"The cat is sleeping $\#$ the chair","under","above"}
    \item 'conjunctions': \\
    \texttt{"She went to the store $\#$ she needed milk.", "because", "although"}
    \item 'temporal understanding': \\
    \texttt{ "It's usually dark outside $\#$ the sun rises", "before", "while"}
    \item 'spatial understanding': \\
    \texttt{"The cat is under the $\#$.", "table", "sky"}
    \item 'quantitative reasoning': \\
    \texttt{"There are 5 apples. If I eat 2, there will be $\#$ left", "3", "4"}
    \item 'emotions': \\
    \texttt{ "When he lost his keys, he was really $\#$.", "frustrated", "excited"}
    \item 'narrative understanding': \\
    \texttt{"After the long journey, the traveler was $\#$ and fell asleep quickly.", "tired", "hungry"}
    \item 'ethics': \\
    \texttt{"Cheating to win a game is $\#$ acceptable", "never", "always"}
\end{itemize}


\end{document}